\documentclass[10pt,letterpaper]{article}

\usepackage{cogsci}

\cogscifinalcopy 

\usepackage{hyperref}
\usepackage{pslatex}
\usepackage{apacite}
\usepackage{float} 

\usepackage[table]{xcolor}
\usepackage{linguex}

\usepackage{soul}
\usepackage[linguistics]{forest}	
\usepackage{url}			
\usepackage{booktabs}		
\usepackage{graphics}		
\usepackage{enumitem}
\usepackage{multirow}
\def\newterm#1{\textit{#1}}
\newcommand{\cites}[1]{\citeauthor{#1}'s \citeyear{#1}}
\definecolor{mygray}{gray}{0.9}

\title{Can neural networks acquire a structural bias from raw linguistic data?}
 
\author{{\large \bf Alex Warstadt (warstadt@nyu.edu)} \\
  Department of Linguistics, New York University \\
  New York, NY 10003 USA
  \AND {\large \bf Samuel R.~Bowman (bowman@nyu.edu)} \\
  Department of Linguistics \& Center for Data Science \& Department of Computer Science, New York University \\
  New York, NY 10003 USA}

\begin{document}

\maketitle

\begin{abstract}
We evaluate whether BERT, a widely used neural network for sentence processing, acquires an inductive bias towards forming structural generalizations through
pretraining on raw data. We conduct four experiments testing its preference for structural vs. linear generalizations in different structure-dependent phenomena. We find that BERT makes a structural generalization in 3 out of 4 empirical domains---subject-auxiliary inversion, reflexive binding, and verb tense detection in embedded clauses---but makes a linear generalization when tested on NPI licensing. We argue that these results are the strongest evidence so far from artificial learners supporting the proposition that a structural bias can be acquired from raw data. If this conclusion is correct, it is tentative evidence that some linguistic universals can be acquired by learners without innate biases. However, the precise implications for human language acquisition are unclear, as humans learn language from significantly less data than BERT.


\end{abstract}

\section{Introduction}

\begin{figure}[t]
    \centering
    \includegraphics[width=\columnwidth]{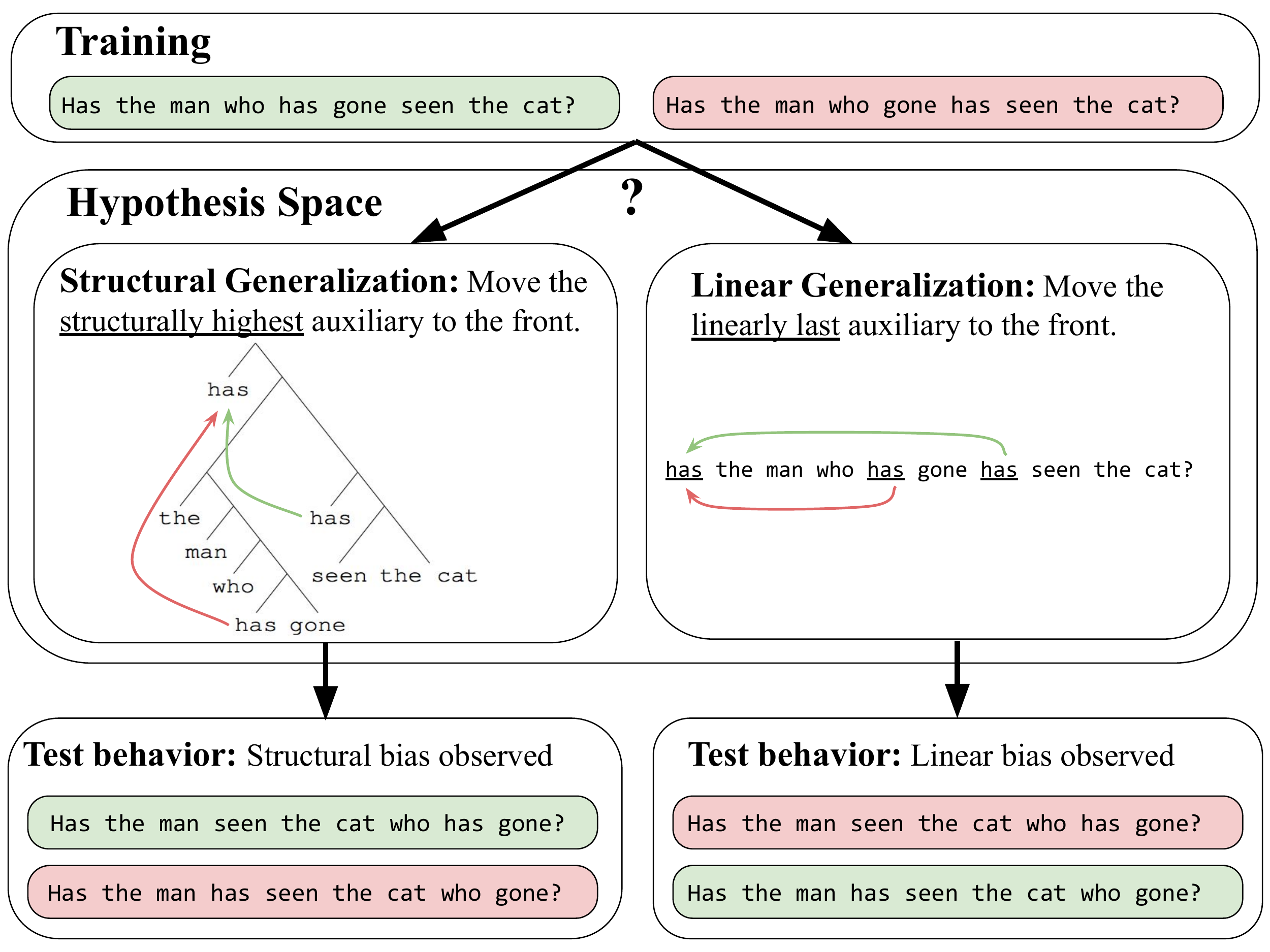}
    \caption{Illustration of the poverty of the stimulus design experiment for subject-auxiliary inversion. Colors correspond to the binary classes for sentences.}
    \label{fig:experiment}
\end{figure}

Humans appear to use structural biases to guide language acquisition. A classic example is the rule for subject-auxiliary inversion: Native English speakers uniformly acquire a rule like the structural generalization in Figure \ref{fig:experiment} that makes reference to hierarchical syntactic structures, despite the fact that the raw linguistic input often supports linear generalizations which are intuitively just as simple \cite{chomsky1965aspects}. Humans are not alone in possessing this inductive bias: Prior investigations have identified some artificial learners with a structural bias by virtue of having a significantly restricted the hypothesis space \cite{perfors2011learnability} or a hierarchically structured architecture that learns from pre-parsed data \cite{mccoy2020does}. 

However, these results cannot tell us whether a learner starting with very weak biases can \emph{acquire} a structural bias merely from exposure to raw linguistic data. While inductive biases are often understood to be unchangeable properties of a learner, this need not be the case. For instance, in one dominant paradigm in natural language processing, \textit{pretraining} on raw data is used to create a general purpose sentence processing model like BERT \cite<Bidirectional Encoder Representations from Transformers;>{devlin2019bert}, which can subsequently be fine-tuned to perform a downstream task. The model's inductive biases with respect to the downstream task may be substantially influenced by the prior knowledge acquired during pretraining.

In this work, we present new experimental evidence that BERT may acquire an inductive bias towards structural generalizations from exposure to raw data alone. 
We conduct four experiments inspired by \citeauthor{mccoy2018revisiting} \citeyear{mccoy2018revisiting,mccoy2020does} to evaluate whether BERT has a structural or linear bias when generalizing about structure-dependent English phenomena. We follow the \emph{poverty of the stimulus} design \cite{wilson2006learning}, outlined in Figure \ref{fig:experiment}. First, we fine-tune BERT to perform a classification task using data intentionally ambiguous between structural and linear generalizations. Then, we probe the inductive biases of the learner by observing how it classifies held-out examples that disambiguate between the generalizations. 
The classification tasks illustrate three structure dependent rules of English grammar regarding subject-auxiliary inversion, reflexive-antecedent agreement, and negative polarity item (NPI) licensing. A fourth task requires the classifcation of sentences based on an arbitrary rule: whether the verb in an embedded clause is in the past tense. The data is generated from templates using the generation tools and lexicon built by \citeA{warstadt2019investigating} and \citeA{warstadt2019blimp}.

The results of these experiments suggest that BERT likely acquires a inductive bias towards structural rules from self-supervised pretraining. BERT generalizes in a way consistent with a structural bias in 3 out of 4 experiments: those involving subject-auxiliary inversion, reflexive binding, and embedded verb tense detection. 
While these experiments leave open several alternative explanations for this generalization behavior, they add to mounting evidence that significant syntactic knowledge, including a structural biases, can be acquired from self-supervised learning on raw data.



\section{Background \& Related Work}

\subsection{Self-supervised Learning and BERT}

Recent advances in machine learning for natural language processing give new reason to believe that low-bias learners can acquire significant grammatical knowledge from raw data. The Transformer neural network architecture \cite{vaswani2017attention} used in models like BERT has very weak biases: It is a universal approximator of the class of sequence transduction functions \cite{yun2019transformers}, and it has been applied effectively in non-linguistic domains such computer vision \cite{parmar2018image} and protein sequence modeling \cite{rives2019biological}. However, rather than training a low bias model from scratch to perform a particular linguistic task, recent results show that it is far more more effective to pretrain a general purpose model on raw data and subsequently fine-tune it on a downstream task \cite<e.g.>{howard2018universal}. 
The implication is that they acquire helpful biases from pretraining. 

Crucially, these models are usually pretrained with only raw data using the technique of \newterm{self-supervised learning}, which circumvents the need for labeled data by using the data itself as the labels. The most common self-supervised tasks for pretraining are the language modeling task, where the objective is predict the next word in a string \cite<e.g.>{peters2018elmo,radford2019language}, and, in the case of BERT, the cloze task where the objective is predict the identity of a masked token anywhere in a string. 

Despite containing no explicit information about grammatical concepts, self-supervised tasks appear to teach neural models significant knowledge of grammar and hierarchical syntax. These models can perform human-like acceptability judgments, which are understood in linguistics as a probe on linguistics competence \cite{schutze1996empirical}. When fine-tuned to perform acceptability judgments, BERT approaches human performance on the Corpus of Linguistic Acceptability \cite<CoLA;>{warstadt2019neural}, a dataset of over 10k example sentences from linguistics publications with Boolean acceptability judgments. Language models can also correctly discriminate minimal pairs for subject-verb agreement \cite{gulordava2019colorless}, wh-dependencies \cite{wilcox2018filler}, and numerous other linguistic phenomena in English \cite{warstadt2019blimp} without any supervised training on acceptability. BERT's internal representations appear to attend to linguistic features such as syntactic category \cite{clark2019does} and contain sufficient information from which to recover a dependency parse for an inputted sentence \cite{hewitt2019structural}. However, it is not known whether BERT is biased towards forming generalizations based on structural features when fine-tuned on structure-dependent phenomena. Indeed, it is possible that BERT could acquire knowledge of hierarchical syntax but still preferentially use surface features to generalize. Our experiments are designed to address this question.

\subsection{Structure Dependence \& the Innateness Hypothesis}

The learnability of structural bias has played a large role in debates about human language acquisition. \citeA{chomsky1965aspects,chomsky1971problems} proposes that humans have an innate bias towards learning structural grammatical rules.
For example, children must learn a general rule for subject-auxiliary inversion primarily from input like \ref{eg:sentence pair}. With this input, a learner could form a structural generalization (front the highest auxiliary in the corresponding declarative) or a linear one (e.g. front the first or last auxiliary), but human learners always choose the former. That is, no human learner of English acquires a linear rule that produces the form in \ref{eg:bad_linear} over \ref{eg:bad_struc}. From such examples, \citeA{chomsky1971problems} concludes that humans have an innate preference for structure-dependent rules. Otherwise it would be difficult to explain how we so consistently avoid deeply un-language-like hypotheses in lieu of significant disconfirming evidence.

\ex.\label{eg:sentence pair}
\a. The cat has gone.
\b. Has the cat gone?

\ex.\label{eg:bad}
\a.\label{eg:bad_linear} *Is the man has seen the cat who going?
\b.\label{eg:bad_struc} Has the man seen the cat who is going?




These examples play a key role in Chomsky's influential argument from the poverty of the stimulus in support of this hypothesis. A version of the argument is given below:\footnote{See \citeA{laurence2001poverty} and \citeA{pullum2002empirical} for a more detailed exposition of this argument.}

\begin{itemize}[leftmargin=*]
\small
    \item[] \textbf{Premise 1}\, Humans form grammatical hypotheses about their native language using either innate biases or data-driven learning.
    \item[] \textbf{Premise 2}\, Human language learners preferentially form structural hypotheses over equally simple linear hypotheses.
    \item[] \textbf{Premise 3}\, The raw linguistic input during language acquisition favors neither the structural nor the linear hypothesis.
    \item[] \textbf{Conclusion}\, Humans' preference for structural generalization is not learned from the raw linguistic input (or any other part of the learner's environment), i.e., it is innate.
\end{itemize}

This conclusion has spurred much fruitful research into the nature of linguistic universals \cite<see e.g.,>{chomsky1981lectures} and the argument has been fleshed out with evidence child language acquisition \cite{crain1987structure,yang2000knowledge}. Nonetheless the argument's validity has been at times been questioned \cite{pullum2002empirical,reali2005uncovering,perfors2011learnability}. If the premises are granted, the reasoning to the conclusion is sound. Premise 1 is a tautology given a suitable definition of data-driven learning. Premise 2 is an robust empirical result of generative linguistics.

Therefore, those arguing against this result have generally taken issue with Premise 3 of the argument, which I refer to henceforth as the \newterm{impoverishment assumption}. For instance, \citeA{pullum2002empirical} conduct a corpus search to show that such sentences that would provide evidence for the structural rule such as \ref{eg:bad_struc} are attested in naturalistic speech. However, \citeA{legate2002empirical} counter that such examples are insufficiently frequent to meaningfully impact learning. More importantly, whether or not crucial evidence is lacking for the acquisition of subject-auxiliary inversion, there are certainly numerous other structural rules which have been proposed for natural languages for which crucial examples are vanishingly rare.

\citeA{reali2005uncovering} articulate a broader criticism of the impoverishment assumption. It is not only sentences of the form in \ref{eg:bad_struc} that count as evidence for the structural hypothesis. Rather, data from all domains of language provide indirect evidence that militates in favor of a structural understanding of grammar. Statistical regularities in language, they suggest, may be sufficient for ``bootstrapping syntax''. Indeed, \citeA{perfors2011learnability} find that Bayesian grammar induction system given a choice between several grammar types will preferentially hypothesize phrase-structural rules for English over a flat grammar when presented with child directed speech. This result suggests that the raw data seen be children does favor the acquisition of structural rules in general, counter to the impoverishment assumption, at least if the hypothesis space is strictly limited. However, their models are not low-bias learners, nor do they discover syntactic representations on their own. Rather, they are presented with several hand-crafted candidate grammars of various types including a generally adequate phrase-structure grammar, as well as the syntactic categories of the input data.





\begin{table*}[t]
    \centering
    \begin{tabular}{lp{0.08\textwidth} p{0.36\textwidth} p{0.36\textwidth}}
    \toprule
         \textbf{Experiment} 
         &\textbf{Set} 
         &\textbf{Acceptability} 
         &\textbf{Unacceptable}  \\\midrule
        
    \multirow{2}{0.09\textwidth}{\textbf{S-Aux-Inv}}
       & \textbf{Training}
       & \cellcolor{mygray} Has the man who is going seen the cat?
       &  Is the man who going has seen the cat?\\
       & \textbf{Test}
       &  Has the man seen the cat who is going?
       & \cellcolor{mygray} Is the man has seen the cat who going?\\\midrule
       
       \multirow{2}{0.08\textwidth}{\textbf{Reflexive}}
       & \textbf{Training}
       & \cellcolor{mygray} The boy that loves himself talks to ladies. 
       & The boy that loves themselves talks to ladies?\\
       & \textbf{Test} 
       & \cellcolor{mygray} The boy that loves ladies talks to himself. 
       & \cellcolor{mygray} The boy that loves ladies talks to themselves.\\\midrule
       
       \multirow{2}{0.08\textwidth}{\textbf{NPI}}
       & \textbf{Training} 
       & \cellcolor{mygray} Kids who saw the cats won't get any dogs. 
       & Kids who saw any cats won't get the dogs.\\
       & \textbf{Test} 
       & \cellcolor{mygray} Kids who won't see any cats get the dogs. 
       & \cellcolor{mygray} Kids who won't see the cats get any dogs. \\\toprule
       && \textbf{Embedded Past} & \textbf{Embedded Present}\\\midrule
       
       \multirow{2}{0.08\textwidth}{\textbf{Tense}}
       & \textbf{Training}
       & \cellcolor{mygray} The critic who sang arias praised a lady. 
       & The critic who sings arias praised a lady. \\
       & \textbf{Test}
       & \cellcolor{mygray} The critic praised a lady who sang arias. 
       & \cellcolor{mygray} The critic praised a lady who sings arias. \\
       \bottomrule
    \end{tabular}
    \caption{Data from the subject-auxiliary inversion experiment. According to the relevant linear generalizations, sentences shaded in gray will belong to the positive class, and sentences in white belong to the negative class.}
    \label{tab:data}
\end{table*}

\subsection{Testing the Biases of Neural Networks}\label{sec:posd}

There have been several prior efforts to test neural networks for a structural bias in the domain of subject-auxiliary inversion \cite{lewis2001learnability,frank2007transformational,mccoy2018revisiting,mccoy2020does}. These studies all adopt some form of the \newterm{poverty of the stimulus design} \cite{wilson2006learning}, an experimental paradigm that probes the inductive biases of learners by training them on data that is ambiguous between several hypotheses, and evaluating them on examples that disambiguate between these hypotheses. For instance, in a series of papers  \citeauthor{mccoy2018revisiting} \citeyear{mccoy2018revisiting,mccoy2020does} train neural networks to generate a polar question from the corresponding declarative, using training and test data similar to our subject-auxiliary inversion paradigm shown in Table \ref{tab:data}. They find that, while 
tree-structured models trained using parsed data make a structural generalization, low-bias models never do so consistently.

However, there is good reason to revisit this question with BERT. \citeauthor{mccoy2018revisiting}~do not evaluate BERT, but rather LSTMs that are pretrained on an auto-encoding task in which the model must reproduce the input sentence verbatim. Furthermore, the pretraining data is not naturally occurring, but generated from a restricted lexicon and small context-free grammar. \citeA{conneau2018cram} have already been shown that auto-encoders are much weaker at learning syntactic features than surface features even with naturalistic training data. Thus, there is little reason to expect that BERT would perform similarly at this task.

\section{Materials \& Methods}

We apply the poverty of the stimulus design to test whether unsupervised pretraining gives BERT a structural bias. We conduct four experiments using semi-automatically generated data illustrating different structural generalizations, including the subject-auxiliary inversion paradigm investigated by \citeA{mccoy2018revisiting,mccoy2020does}. Each experiment consists of a training phase in which BERT is fine-tuned to classify sentences from an impoverished paradigm consistent with both a structural hypothesis and a linear hypothesis. Then, the classifier is evaluated on the full paradigm which disambiguates between the two hypotheses. If the classifier makes the structural generalization, this is evidence that BERT has learned a structural bias from pretraining on raw data. 



\subsection{Data and Tasks}

Examples from the four experimental datasets are shown in Table \ref{tab:data}. Each dataset is associated with a binary classification task. In the subject-auxiliary inversion, reflexive, and NPI datasets, the classes correspond to grammatical acceptability of the sentence. In the tense dataset, the classes correspond to whether or not the embedded verb appears in the past tense. 

The tense detection task enables us to draw additional conclusions not possible with the acceptability task alone. Since BERT appears to acquire some knowledge of acceptability from pretraining, it might converge on the structural generalization on the acceptability task not by forming a new structural generalization from ambiguous training data, but by accessing an implicit structural rule acquired during pretraining. In the tense detection task there is no reason to expect BERT has acquired the structural generalization during pretraining, and the structural and linear hypotheses are equally arbitrary. Thus, this paradigm tests whether BERT has a structural bias when forming completely novel generalizations.

All datasets and generation scripts are available at the following link:
\href{https://github.com/alexwarstadt/data_generation/tree/structural_bias_cogsci}{\url{https://github.com/alexwarstadt/data_generation/tree/structural_bias_cogsci}}.



\paragraph{Subject-Auxiliary Inversion} In the subject-auxiliary inversion dataset, the structural and linear hypotheses are defined in terms of where the auxiliary at the front of the sentence has moved from. Each sentence contains two clauses---a main clause and an embedded relative clause---each with an auxiliary verb. In the training examples the embedded auxiliary precedes the main auxiliary because the relative clause modifies the subject. In the test examples the embedded auxiliary follows the main auxiliary because the relative clause modifies the object. Therefore, a linear generalization which targets the last auxiliary will give the same result for the training as a structural generalization that targets the main auxiliary, but the two generalizations diverge for the test examples.

\paragraph{Reflexive Binding} In the reflexive dataset, adapted from \citeA{marvin2018targeted}, the structural and linear hypotheses depend on the relation between the reflexive pronoun (e.g. \emph{himself}) and a \newterm{binder} noun phrase that agrees with it in person and number (e.g. \emph{the boy}). To a first approximation, the structural c-command relation must hold between the binder and reflexive in order for the sentence to be acceptable \cite{chomsky1981lectures}. However, in each training examples, the binder also precedes the reflexive only in the acceptable sentences. These generalizations diverge in the test examples, in which there is an unacceptable sentence where the binder precedes, but does not c-command, the reflexive.

\paragraph{NPI Licensing} In the NPI dataset, also adapted from \citeA{marvin2018targeted}, the structural and linear hypotheses depend on the relation between a negative polarity item (NPI; e.g., \emph{any}) and negation. Loosely speaking, negation must c-command the NPI in order for the sentence to be acceptable. The hypotheses are the same (\emph{mutatis mutandis}) as those for the reflexive dataset.

\paragraph{Tense} In the tense dataset, each sentence has a main verb and an embedded verb, and the classes correspond to whether the embedded verb has past tense inflection. A related objective is used to evaluate structural knowledge in pretrained models by \citeA{shi2016syntax} and \citeA{conneau2018cram}. As with the subject-auxiliary inversion dataset, the embedded verb precedes the main verb in the training examples, and follows it in the test examples. Therefore, structural and linear generalizations about the position of the verb learned from the training data will diverge for the test data.

\paragraph{Data Generation} The data for the experiments are automatically generated using a method similar to that adopted by \citeA{ettinger2016probing} and \citeA{marvin2018targeted}. Sentences are generated from templates describable by a simple context-free grammar. Lexical items are sampled from a hand-crafted vocabulary of over 1000 items labeled with over 30 features required for morphological, syntactic, and semantic well-formedness. Each generated dataset consists of training, development, and test sets each with 10k examples. Examples with similar lexical content are generated in sets of 4, as in Table \ref{tab:data}. Within a set of 4, the training items form a minimal pair, as do the test items.

\subsection{Methods}

In each experiment, we train 20 random restarts of classifiers on top of BERT using the labeled training items. Following \citeA{devlin2019bert}, we fine-tune BERT itself in addition to the classifier layer during training. We use Huggingface's \cite{wolf2019huggingface} implementation of BERT-Large in PyTorch \cite{paszke2017automatic}, and carry out fine-tuning in \texttt{jiant} \cite{wang2019jiant}. All 20 random restarts for each experiment use identical hyperparameters. The only difference between them is the random seed used to initialize the classifier weights. The models use a learning rate of 2e-5, a dropout rate of $0.2$, and a batch size of $16$. Training is carried out for $4$ epochs, or until $5$ evaluations occur without any improvement in development accuracy. During training, the model is evaluated on the development set after $10$ batches. These hyperparameters are selected based on an exploratory grid search using the recommended hyperparamer ranges suggested by \cite{devlin2019bert}.
The hyperparameters selected for the experiments consistently led to the best or near-best development accuracy for each of the datasets.

\begin{figure}
    \centering
    \includegraphics[width=\columnwidth]{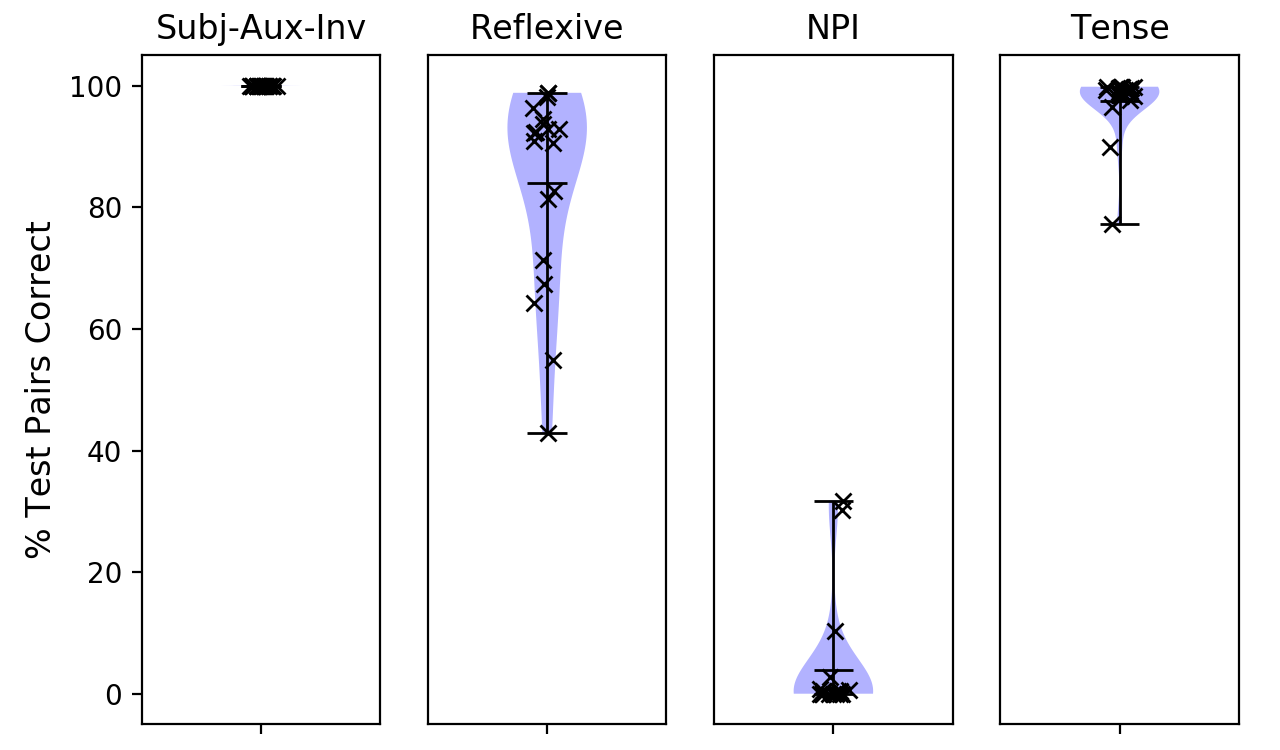}
    \caption{Test results for 20 random restarts of the 4 experiments. ``\% Test Pairs Correct'' is the percentage of minimal pairs from the test templates correctly classified. Individual runs are plotted as \emph{x}'s with random horizontal jitter.}
    \label{fig:results}
\end{figure}

\section{Results}

The results for the experiments in Figure \ref{fig:results} show that BERT has an overwhelming tendency to generalize in a way consistent with structural generalization in the subject-auxiliary inversion, reflexive, and tense settings, but not in the NPI setting. This plot shows the proportion of test minimal \emph{pairs} classified correctly, rather than the proportion of test items classified correctly. Because there are four ways to classify a pair of items, a totally random classifier would have an expected accuracy of 25\% on this metric. In each experiment, minimal pair accuracy on held-out examples from the training templates is over 90\%, indicating that the models robustly learned a generalization consistent with the training data.

On the subject-auxiliary inversion task, BERT classifiers appear to make structural generalization with very high consistency and near perfect accuracy. All 20 classifiers exceed 99\% minimal pair accuracy.

The reflexive classifiers also appear to make the structural generalization in most cases, but are less consistent. The median minimal pair accuracy is over 92\% and the maximum exceeds 99\%. However, there is a portion of classifiers that classify the test minimal pairs correctly only about half the time, despite achieving high performance on the training examples. The other half of the time, they tend to classify the test examples in a way consistent with the linear hypothesis. However none of these classifiers \emph{systematically} makes predictions consistent with the linear generalization. 

The tense classifiers make a structural generalization very consistently, with a median minimal pair accuracy over 99\%. There are two outliers where accuracy is between 75\%-90\%.

Finally, the results from the NPI experiment are the exceptional case where we do not observe behavior consistent with a structural generalization. The median minimal pair accuracy is effectively 0\%, and the maximum is barely above chance at 32\%. While every model classifier is able to classify the training pairs perfectly, only 6/20 consistently classify the test pairs in the same way. All of these classifiers learn a generalization that is different from the hypothesized linear generalization shown in Table \ref{tab:data}. We find that their performance is consistent with grouping together all and only sentences with an NPI towards the end of the sentence.

\section{Discussion}\label{sec:discussion}

These results suggest that it is likely that BERT does acquire some form of a structural inductive bias from self-supervised pretraining, at least outside of the NPI domain.
They point more strongly in this direction than earlier results by \citeA{mccoy2018revisiting,mccoy2020does}.
If this interpretation is correct, it would cast some doubt on the impoverishment assumption from \cites{chomsky1965aspects} argument from the poverty of the stimulus by showing that raw data does contain overwhelming evidence that language is hierarchical. If some learner does not require innate bias to discover the utility of preferring structural rules over linear ones, it stands to reason humans may not either. On the other hand, our results are consistent with other interpretations, and so we caution against leaping to this strong conclusion, at least without further evidence.

\paragraph{The NPI Results}

First, in the NPI domain, BERT does not show a structural bias. However, it does not immediately follow that BERT does not have a structural bias at all. By virtue of the paradigm's design, the classes that the model converged on are consistent with both a linear and a structural position. As mentioned above, 6/20 classifiers classified the test items systematically, though not in the way predicted in Table \ref{tab:data}. Instead, they grouped the top left and bottom right sentences in one class, and the top right and bottom left sentences in the other. The sentences in the first class might be characterized as all sentences with an NPI at the end of the sentence (a linear generalization), or all sentences with an NPI in the main clause (a structural generalization). Additional experiments are needed to determine which of these outcomes has occurred.

Furthermore, humans do not necessarily show a structural bias in processing similar examples. The unacceptable test items in the NPI paradigm, where the negation precedes the NPI, are known as \newterm{NPI illusions}, because they can spuriously appear to be acceptable to humans, and pattern with grammatical sentences in self-paced reading and ERP experiments \cite{xiang2009illusory}. 
Thus, NPIs may in retrospect not be the clearest example of humans' structural bias.

\paragraph{Limitations of the Poverty of the Stimulus Design} 
Some doubts remain even in the domains where BERT appears to show a structural generalization. Some surface features could accidentally give the same predictions as the structural generalization on the test data. For instance, in the subject-auxiliary inversion data in Table \ref{tab:data}, a classifier could coincidentally identify the acceptable examples by learning to identify a string with a relativizer adjacent to an auxiliary (e.g. \emph{who is}). In fact, we control for this particular confound by generating acceptable examples where a finite verb follows the relativizer (e.g. \emph{Has the man who went seen the cat?}).

However, there are likely other surface generalizations that are consistent with the results. This problem can be addressed by training and evaluating on data that contradict these generalizations, but alternative convergent hypotheses cannot be eliminated entirely. This is a fundamental limitation of the poverty of the stimulus design: It is not possible to determine that BERT is adopting any particular generalization. 

That said, if we continue to find convergent from multiple unrelated domains, Occam's Razor tells us that we should conclude BERT has a structural bias. Given the large number of conceivable surface generalizations, let us assume that an arbitrary generalization is equally likely to support any of the four classification behaviors for the test minimal pair. It follows that if the classifier does make some surface generalization, there is a 1 in 4 chance for each experiment that it would accidentally align with the structural generalization. Then the probability that this chance alignment would occur in at least 3 out of 4 domains is about 5\%.


This worry could be alleviated further if it could be shown that baseline models without unsupervised pretraining tend to make the linear generalization on these datasets. These experiments will have to be included in future work. However, at present, the results of \cites{mccoy2018revisiting} experiments can be used as a proxy. As described in Section \ref{sec:posd}, these experiments test the ability of sentence encoders without substantial unsupervised pretraining to generalize from a paradigm resembling the polar question data in my experiments. In 5 out of 6 of the model architectures they tested, the linear generalization was preferred. While the task in their experiment is different the acceptability judgment task in the present work, based on this finding it seems that sequence models without substantial unsupervised pretraining are likely to prefer the linear generalization in the polar question domain.

\paragraph{Conclusion}

This work presents new evidence that highlights the possibility that language learners could \textit{acquire} a structural inductive bias from statistical regularities in raw linguistic data. In particular, we find the most comprehensive evidence to date (to our knowledge) of a low-bias learner demonstrating a structural bias acquired through unsupervised learning on raw data. However, evidence from other empirical domains is needed to fully evaluate this conclusion. 

Future work should draw more direct connections between neural networks and human language learners. BERT is trained on data from domains far outside the input to human learners, and in much greater quantities. Indeed, the quantity and quality of data are two other pillars of \cites{chomsky1965aspects} argument from the poverty of the stimulus. Therefore, it is essential to replicate these experiments with models trained on less data, which bears greater resemblance to the input to a child. Furthermore, \cite{chomsky1965aspects} observes that human languages generally lack linear rules. If neural networks can acquire human-like biases, they should also struggle to form certain kinds linear generalizations. As techniques for machine language learning and self-supervised pretraining continue to advance, we expect to learn more about which linguistic universals are and are not learnable from data.



\section*{Acknowledgments}

We thank Chris Barker, Chris Collins, Stephanie Harves, Brenden Lake, Tal Linzen, Alec Marantz, Tom McCoy, and the audience at NYU's Syntax Brown Bag for helpful feedback. This material is based on work supported by the National Science Foundation under Grant No. 1850208. 
Any opinions, findings, and conclusions or recommendations expressed in this material are those of the author(s) and do not necessarily reflect the views of the National Science Foundation. 
This project has also benefited from support to SB by Eric and Wendy Schmidt (made by recommendation of the Schmidt Futures program), by Samsung Research (under the project Improving Deep Learning using Latent Structure), by Intuit, Inc., and by NVIDIA Corporation (with the donation of a Titan V GPU).

\bibliographystyle{apacite}

\setlength{\bibleftmargin}{.125in}
\setlength{\bibindent}{-\bibleftmargin}

\bibliography{CogSci_Template}

\end{document}